\documentclass[runningheads]{llncs}
\usepackage{comment}
\usepackage{graphicx}
\usepackage{adjustbox}
\usepackage{multirow}
\usepackage{marvosym}
\usepackage{float}
\usepackage{subcaption}
\usepackage{booktabs} 
\usepackage{amsmath}
\usepackage{makecell}
\usepackage{array}
\usepackage{hyperref}
\hypersetup{hidelinks,
	colorlinks=true,
        linkcolor=black,
        filecolor=black,      
        urlcolor=magenta,
	pdfstartview=Fit,
	breaklinks=true}

\begin{document}
\title{Perceptual Quality Assessment of Omnidirectional Audio-visual Signals}

\def\CICAISubNumber{241}  
\author{Xilei Zhu \and
Huiyu Duan \orcidID{0000-0002-6519-4067} \and
Yuqin Cao \and
Yuxin Zhu \and
Yucheng Zhu \and
Jing Liu \and
Li Chen \and
Xiongkuo Min \and
Guangtao Zhai\textsuperscript{(\Letter)}
}
\authorrunning{Zhu et al.}
\institute{Institute of Image Communication and Network Engineering, \\Shanghai Jiao Tong University \\
\email{\{xilei\_zhu,~huiyuduan,~caoyuqin,~rye2000,~zyc420,~hilichen,\\~minxiongkuo,~zhaiguangtao\}@sjtu.edu.cn,~jliu\_tju@tju.edu.cn}\\}

\maketitle              

\begin{abstract}
Omnidirectional videos (ODVs) play an increasingly important role in the application fields of medical, education, advertising, tourism, \textit{etc}. Assessing the quality of ODVs is significant for service-providers to improve the user's Quality of Experience (QoE).
However, most existing quality assessment studies for ODVs only focus on the visual distortions of videos, while ignoring that the overall QoE also depends on the accompanying audio signals.
In this paper, we first establish a large-scale audio-visual quality assessment dataset for omnidirectional videos, which includes 375 distorted omnidirectional audio-visual (A/V) sequences generated from 15 high-quality pristine omnidirectional A/V contents, and the corresponding perceptual audio-visual quality scores.
Then, we design three baseline methods for full-reference omnidirectional audio-visual quality assessment (OAVQA), which combine existing state-of-the-art single-mode audio and video QA models via multimodal fusion strategies. We validate the effectiveness of the A/V multimodal fusion method for OAVQA on our dataset, which provides a new benchmark for omnidirectional QoE evaluation. Our dataset is available at \href{https://github.com/iamazxl/OAVQA}{https://github.com/iamazxl/OAVQA}.
\vspace{-8pt}
\vspace{-8pt}
\end{abstract}
\section{Introduction}
\vspace{-6pt}
Virtual Reality (VR) has attracted substantial attention from industry and research communities due to its ability to provide users with a stereoscopic and immersive experience through Head-Mounted Displays (HMDs) \cite{duan2018perceptual,duan2022saliency}. 
Omnidirectional Videos (ODVs), \textit{a.k.a}, $360^{\circ}$ videos, panoramic videos or spherical videos, have emerged as a significant form of VR content. By using VR HMDs and adjusting their head orientation, users can explore the audio-visual content in any direction. This immersive experience of simulating real-world scenes has contributed to the popularity of ODVs in various application fields, including medical, education, advertising, tourism, \textit{etc}.


Compared to traditional videos, ultra high-definition ODVs contain more scene information and multi-channel audio information, which results in a doubling of ODV data volume. Due to the huge amount of data, playback stucking and quality switching caused by network delays and fluctuations usually occur during video transmission, which leads to the degradation of ODVs quality, and further affects the QoE of ODVs. Moreover, ODVs may also suffer from the distortions introduced during the process of capturing or displaying, which further decreases the QoE.
Therefore, to provide users with a smooth viewing experience, it is important to monitor the quality of ODVs during the procedure of shooting, codec, transmission, \textit{etc.}, and perform optimization accordingly.



 
In the past few decades, many objective quality assessment methods have been proposed for traditional plane videos~\cite{sheikh2006image,li2016toward}, and some recent works have also explored the problem of audio-visual video quality assessment~\cite{min2020study}. Recently, with the popularity of VR, many studies have explored the problem of omnidirectional image quality assessment \cite{duan2023attentive,sun2019mc360iqa} and omnidirectional video quality assessment~\cite{fela2021perceptual}. However, most omnidirectional video quality assessment research only focuses on the single-mode signal, \textit{i.e.}, visual information, few works have investigated the multimodal quality assessment of ODVs incorporating audio information. As an important part of ODVs, spatial audio may strongly influence the human perceptual quality, thus it is necessary to conduct in-depth research on the audio-visual quality assessment of the omnidirectional videos.

In this paper, we make three contributions to the omnidirectional audio-visual quality assessment (OAVQA) field. Firstly, we construct a large-scale omnidirectional audio-visual quality assessment dataset to solve the poverty problem of the corresponding dataset. We first collected 15 high-quality reference omnidirectional audio-visual (A/V) content, and generated 375 distorted ODVs degraded from them. Subsequently, 22 subjects were recruited to participate in the subjective quality assessment experiment, and the audio-visual quality ratings of the reference and distorted videos were collected.
Secondly, we design three baseline methods for  full-reference omnidirectional AVQA. The baseline models first utilize the existing state-of-the-art audio and video single-mode quality assessment methods to predict the audio quality and video quality of ODVs, respectively, then utilize different multimodal fusion strategies to fuse A/V prediction results and obtain the overall quality results of the ODVs. Thirdly, we compare and analyze the prediction performance of these models on our dataset, and establish a new benchmark for future studies on OAVQA.


\vspace{-8pt}
\section{Related Work}
\vspace{-8pt}
\subsection{Omnidirectional Video Quality Assessment Dataset}
\vspace{-6pt}
Table \ref{dataset} provides an overview of several existing omnidirectional video quality assessment datasets. 
It can be observed that most of the existing ODV quality assessment datasets lack spatial audio information, and mainly focus on visual distortions, while audio distortions are rarely been considered.

\begin{table}[t]
\centering
\caption{An overview of omnidirectional video quality assessment datasets. ``Mute" means mute audio and ``ambisonics" indicates spatial audio. SI and TI represent spatial information and temporal information respectively. QP indicates quantization parameter and CRF means constant rate factor, which is used to control the video bitrate.}
\label{dataset}
\begin{adjustbox}{width=\textwidth}
\begin{tabular}{|c|c|c|c|c|}
\hline
Dataset  &  Video Num & Audio & Distortion Type & QoE \\
\hline
Schatz \textit{et al.}~\cite{7965657} & 10  & Mute & Stalling & MOS(1$\sim$5) \\
Meng \textit{et al.}~\cite{9349097} & 774 & Mute & Frame size, Frame rate, Quantization stepsize, Resolutions & MOS(1$\sim$10)\\
Fei \textit{et al.}~\cite{fei2019qoe}      & 468   & Mute   & Bandwidth, Packet loss, Latency, Presence   & MOS(1$\sim$5)  \\ 
Anwar \textit{et al.}~\cite{anwar2020measuring}      & 208   & Mute   & Bitrate, Stalling & MOS \\ 
Fan \textit{et al.}~\cite{fan2022modeling}      & 48   & Mute   & Bitrate, Gender, Presence, TI, SI   & MOS(0$\sim$9)  \\ 
IVQAD~\cite{duan2017ivqad}      & 150   & Mute   &  Bitrate, Frame rate, Resolution & MOS(1$\sim$5)  \\ 

VQA-ODV~\cite{li2018bridge}      & 600   & Mute   & QP, Projection format   & DMOS($0\sim60$)  \\ 
Fela \textit{et al.}~\cite{fela2022perceptual}      & 576   & Ambisonics   & QP, Resolution, Audio bitrate   & MOS($0\sim100$)  \\ 
\textbf{Ours} & \textbf{390} & \textbf{Ambisonics} & $\begin{array}{c}
\textbf{Audio bitrate, CRF, Resolution,} \\
\textbf{Noise, Blur, Stucking}
\end{array}$
&  \textbf{MOS(1$\sim$10)} \\ 
\hline
\end{tabular}
\end{adjustbox}\vspace{-20pt}
\end{table}

\vspace{-10pt}
\subsection{Quality Assessment Models}
\vspace{-6pt}

\subsubsection{Omnidirectional video quality assessment.} 
As a common storage format of ODVs, ERP projection has severe mapping stretches near the poles.
In order to solve this problem, Yu \textit{et al.} \cite{yu2015framework} proposed a spherical PSNR scheme (S-PSNR), which is based on a set of uniform sampling points on the spherical surface, the corresponding position on the mapping plane is calculated by different mapping formulas. 
Sun \textit{et al.} proposed the Weighted to Spherically uniform PSNR (WS-PSNR) \cite{sun2017weighted}, which is directly performed in the original format and combined with different stretching weights according to different mapping methods. 
Anwar \textit{et al.} \cite{anwar2020measuring} established an ODVs quality assessment model using the Bayesian inference method, and evaluated the impact of buffering on users' perceptual quality at different bitrates. 
Fan \textit{et al.} \cite{fan2022modeling} established an ODVs dataset that contains various distortions such as compression distortion and quality switching, and then used machine learning methods to establish VQA models. 
\vspace{-15pt}

\subsubsection{Omnidirectional audio-visual quality assessment.}
As an important part of ODVs, the influence of spatial audio on perceptual quality has rarely been studied. 
Zhang \textit{et al.} \cite{8551522} presented a quality assessment methodology for audio-visual multimedia in virtual reality environment. They presented a panoramic audio-visual dataset and the quality factors which represent different distortions were applied as the input to neural network.  
Fela \textit{et al.} \cite{fela2020towards} utilized PSNR and its variants designed for ODVs, \textit{i.e.}, WS-PSNR, CPP-PSNR and S-PSNR~\cite{yu2015framework,Zakharchenko2016,sun2017weighted}, as the quality scores and studied the perceptual audio-visual quality prediction based on the fusion of these scores \cite{fela2021perceptual}.
Four machine learning models including multiple linear regression, decision tree, random forest, and support vector machine (SVM), were tested. 
\vspace{-6pt}
\section{Omnidirectional Audio-visual Quality Assessment Dataset (OAVQAD)}
\vspace{-8pt}
\subsection{Reference and Distorted Contents}
\begin{figure}[t]
    \centering
    \vspace{-3pt}
    \includegraphics[width=0.98\textwidth]{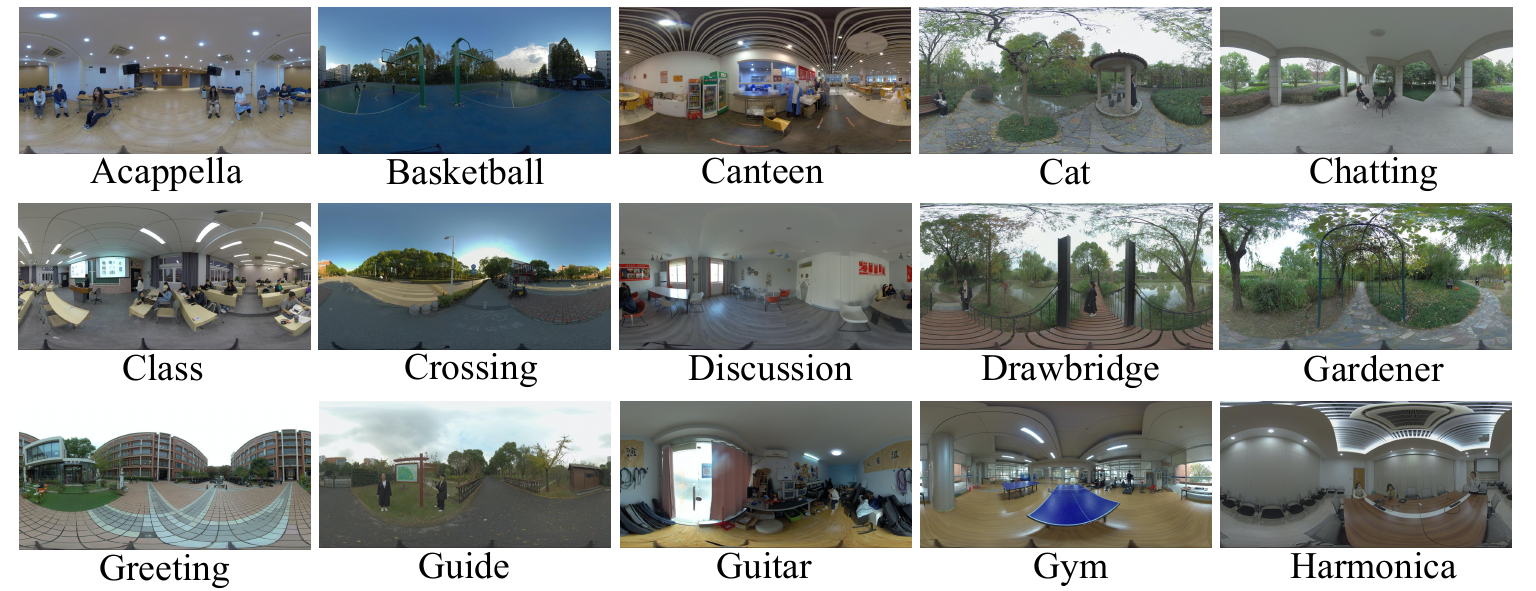}
    \vspace{-5pt}
    \caption{EPR format previews of 15 reference ODVs used in our OAVQAD.}
    \label{25ERP}
    \vspace{-17pt}
\end{figure}

\vspace{-8pt}
We first captured 162 different ODVs with different scenes using a professional VR camera Insta360 Pro2. 
Then, we selected 15 high-quality ODVs from the collected ODVs as the reference videos in our OAVQAD. We utilized FFmpeg to clip the duration of the selected ODVs to 6s.
Each ODV has a resolution of 8K (7680×3840) in equirectangular projection (ERP) format with a frame rate of 29.97 fps. All ODVs contain first order ambisonics (FOA) with 48,000 Hz audio sampling rate and four audio channels. The audio and video formats are shown in Table \ref{parameters}. The ODV contents include acappella chorus, shopping, guitar playing, restaurant ordering, \textit{etc}. Fig.\ref{25ERP} shows the ERP format previews of the selected 15 reference ODVs.


\begin{table}[t]
\centering
\caption{Omnidirectional audio and video format parameters.}
\label{parameters}
\begin{adjustbox}{width=0.9\textwidth}
\begin{tabular}{cccccccc}
\toprule
    & Resolution  & Frame rate  & Bitrate  & Format  & ~Bit depth~ & Duration  & Encoding \\
\midrule
Video & $8K$ & $29.97$fps & $144$Mbps & YUV$420$ & $8$bit & $6$s & H.265 \\
Audio & - & - & $3072$Kbps & FOA & $16$bit & $6$s  & AAC-LC \\
\bottomrule
\end{tabular}
\end{adjustbox}
\vspace{-18pt}
\end{table}

We utilized advanced audio coding (AAC) as the audio encoding method provided by FFmpeg 4.4, and used constant bitrate (CBR) mode to set the audio bitrate to 64Kbps, 32Kbps and 16Kbps, respectively, thereby generating three levels of perceptually well-separated audio compression distortion. Then, we chose HEVC as the video encoding method provided by FFmpeg libx265 encoder, and for each source video we applied 3 different compression levels, \textit{i.e.,} 32, 37 and 42 in constant rate factor (CRF) mode. Besides, we also set the video resolution to three levels including 4K (3840×1920), 2K (1920×960), 1K (1080×540). Moreover, in order to adapt to a wider range of application scenarios, we further introduced more abundant distortion types and added three types of distortions \cite{duan2023masked,duan2022develop} including noise, blur, and stucking, and generated distorted ODVs with various levels of these distortions. To summarize, we applied 25 distortion conditions to 15 reference ODVs, resulting in a total of 375 (15 $\times$ 25) distorted ODVs.

\vspace{-14pt}
\subsection{Subjective Experiment Methodology}
\vspace{-5pt}
\subsubsection{Experiment Apparatus.} 
\begin{figure}[t] 
\center{\includegraphics[width=0.55\textwidth]  {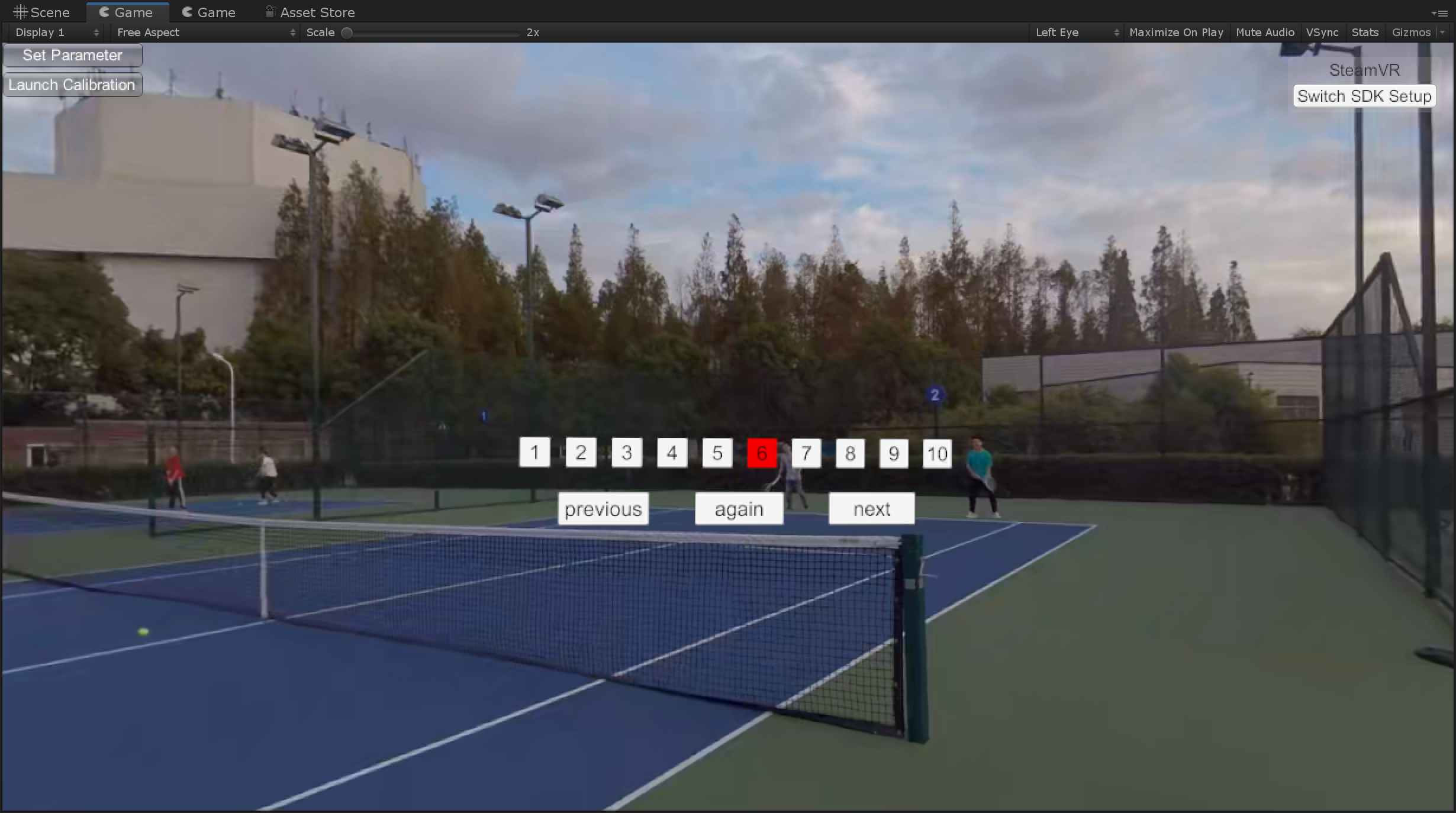}}
\vspace{-5pt}
\caption{Demonstration of the subjective experiment interface based on the Unity platform.}
\label{score}
\vspace{-12pt}
\end{figure}

Since the subjective experiment was needed to be conducted in a VR immersive environment, we used HTC Vive Pro Eye as the HMD to demonstrate ODVs and collect subjective quality ratings. 
The subjective experiment platform used to play 8K ODVs and perform scoring interaction was build based on Unity 1.1.0 as shown in Fig.\ref{score}.
\vspace{-8pt}
\vspace{-8pt}
\subsubsection{Experiment Procedure.} 
The subjective experiment was conducted in a subjective study room in a university. A total of 22 subjects (14 males and 8 females) were invited to participate in the subjective experiment. The subjects were between 20 and 28 years old (mean 22.62, variance 5.23) and were all graduate and undergraduate students. All subjects had normal or corrected-to-normal vision and hearing. 
In the experiment, subjects firstly received the guidance on the use of VR equipment, including HMD and controllers. Then a training session was performed for the subjects, making them be familiarized with the user interface as well as the general range and types of distortions. In the testing session, subjects watched $390$ ODVs and gave perceptual scores of the overall A/V quality. 
 The order of the test videos was random for each subject to avoid bias.

\vspace{-12pt}
\subsection{Subjective Data Processing and Analysis}
\vspace{-4pt}
\begin{figure}[!t] 
\centering
\includegraphics[width=0.65\textwidth, keepaspectratio]  {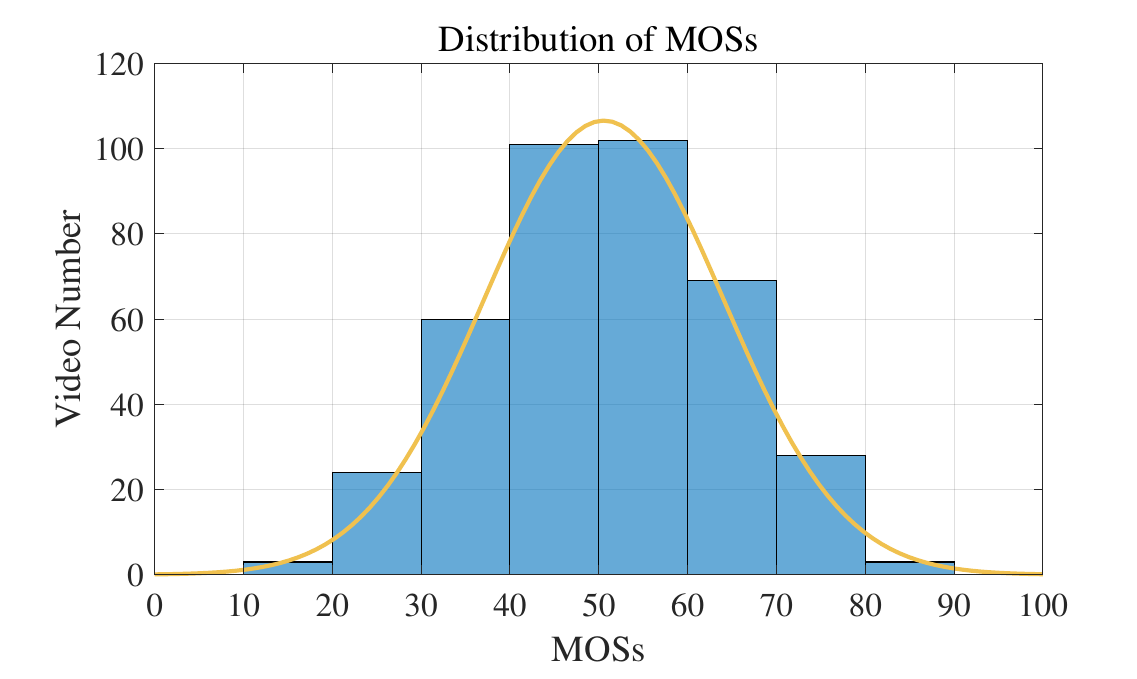}
\vspace{-5pt}
\caption{Histogram of MOS distribution in the database.}
\label{MOS}\vspace{-20pt}
\end{figure}
We followed the subjective data processing method recommended by ITU \cite{duan2022confusing,duan2022augmented} to perform the outlier detection and subject rejection. None of the 22 subjects was identified as an outlier and eliminated. We normalized the raw scores of subjects to Z-scores ranging between 0 and 100 and calculated the mean of Z-scores to obtain the final mean opinion scores (MOSs), which are formulated as follows:
\begin{equation}
    z_{i j}  =\frac{r_{i j}-\mu_i}{\sigma_i}, \quad z_{i j}^{\prime}=\frac{100\left(z_{i j}+3\right)}{6},
\end{equation}
\vspace{-14pt}
\begin{equation}
    \text{MOS}_j  =\frac{1}{N} \sum_{i=1}^N z_{i j}^{\prime},
\end{equation}

where $r_{ij}$ is the original score of the $i$-th subject on the $j$-th sequence, $\mu_i$ and $\sigma_i$ are the mean rating and the standard deviation given by subject $i$, $N$ is the total number of subjects. Fig.\ref{MOS} draws the histogram of MOS distribution over the entire database, indicating that the perceptual quality scores are widely distributed in the $\left[0,100\right]$ interval, basically covering every score segment, and generally showing a normal distribution. It also manifests that the perceptual quality distribution conforms to our expectations and the distortions setting is quite reasonable.


\vspace{-6pt}
\section{Objective Omnidirectional Audio-Visual Quality Assessment}
\vspace{-6pt}
\subsection{Single-mode Models}
\vspace{-4pt}
Many video and audio quality assessment methods have been proposed  separately in previous studies. These quality assessment algorithms, only predict quality of single-modal audio or video signals, can be called as single-mode quality assessment methods. We first utilize the existing state-of-the-art single-mode quality assessment methods to predict the omnidirectional video and audio quality, respectively. 
Since both the single-mode AQA and VQA prediction scores can characterize one aspect of the distortion severity of the distorted video, it is reasonable to directly use the single-mode models to predict the overall audio-visual quality score of the ODVs.

 The well-known single-mode assessment models adopted in this paper are introduced as follows:
\vspace{-5pt}
\begin{itemize}
    \item \textbf{Video}: VMAF~\cite{li2016toward}, SSIM~\cite{wang2004image}, MS-SSIM~\cite{wang2003multiscale}, VIFP~\cite{sheikh2006image}, FSIM~\cite{zhang2011fsim}, 

    \qquad \qquad  GMSD~\cite{xue2013gradient}, WS-PSNR~\cite{sun2017weighted}, CPP-PSNR~\cite{Zakharchenko2016}, S-PSNR~\cite{yu2015framework}.
    
    \item \textbf{Audio}: PEAQ~\cite{thiede2000peaq}, STOI~\cite{taal2011algorithm}, VISQOL~\cite{hines2015visqolaudio}, LLR~\cite{hu2007evaluation}, SNR~\cite{hu2007evaluation}, segSNR~\cite{hansen1998effective}.
\end{itemize}



\vspace{-16pt}
\subsection{Weighted-Product Fusion}
\vspace{-4pt}
A single-mode audio/visual quality assessment metric can only characterize one quality aspect thus cannot fully represent the overall subjective perceptual quality of an ODV. Therefore, it is important to use appropriate multimodal feature fusion method to predict the A/V quality of ODVs.
The simplest fusion method is to directly multiply the quality scores of a VQA model and an AQA model as the overall quality score of ODVs. 

However, for human audio-visual perception, video and audio quality often occupy different importance in ODVs, and people may pay more attention to visual quality. The weighted product can balance the influence of different modalities by assigning different weights to each of them, so the weighted product is a better choice for score fusion compared to the direct multiplication method.
The weighted product can be formulated as
\begin{equation}
\vspace{-4pt}
Q_{a v}=\hat{Q}_v^w \cdot \hat{Q}_a^{1-w},
\end{equation}
where $\hat{Q}_a$ and $\hat{Q}_v$ are normalized score of the audio and video, $w$ and $1-w$ represent the weights of video and audio quality respectively, $0 \leq w \leq 1$. $\hat{Q}_a$ and $\hat{Q}_v$  
are calculated by $\hat{Q}_a=\frac{Q_a-Q_{a_{\min }}}{Q_{a_{\max }}-Q_{a_{\min }}}$ and $\hat{Q}_v=\frac{Q_v-Q_{v_{\min }}}{Q_{v_{\max }}-Q_{v_{\min }}}$, where $Q_{a_{\min }}$, $Q_{a_{\max }}$, $Q_{v_{\min }}$ and $Q_{v_{\max }}$ bound $Q_a$ and $Q_v$ respectively. The optimal weights depend on the used single-mode A/V quality evaluation models and we vary the weight from $0$ to $1$ with $0.05$ step increment to find the optimal weight $w$.
Since the score ranges of the video and audio quality assessment models may be different, the multiplication method can only be performed after they are appropriately scaled or normalized. 

\vspace{-8pt}
\subsection{Support Vector Regression Fusion}
\vspace{-4pt}
Since Support Vector Regression (SVR) is a commonly used machine learning algorithm for establishing nonlinear relationships between inputs and outputs, we also utilize
  the SVR method to integrate the quality prediction scores of single-mode models
  \vspace{-3pt}
\begin{equation}
\vspace{-1pt}
Q_{a v}=\textit{SVR}(Q_v, Q_a),
\vspace{-1pt}
\end{equation}
where $Q_v$ and $Q_a$ represent the quality prediction scores of video and audio, respectively, and $Q_{av}$ denotes the fused A/V quality scores. In this case, SVR uses the single-mode quality scores predicted by traditional AQA and VQA algorithms respectively as the inputs, and the quality score (\textit{i.e.,} MOS) as the labels for regression function training. 

\begin{table}[t]
\vspace{-10pt}
\centering
\caption{Video and audio quality prediction algorithms and their corresponding feature types.}

\label{Features}
\begin{adjustbox}{width=0.92\textwidth}
\begin{tabular}{clcc}
\toprule Category~~~ & Models & Feature~ & \makecell[c]{Decomposed features.}  \\
\hline \multirow{9}{*}{ Video~ } & VMAF \cite{li2016toward} & 6 & 4 scales of VIF, detail loss, motion \\
& SSIM \cite{wang2004image} & 2 & Luminance similarity, contrast and structural similarity \\
& MS-SSIM \cite{wang2003multiscale} & 6 & Luminance similarity, 5 scales of contrast and structural similarity \\
& VIFP \cite{sheikh2006image} & 4 & 4 scales of VIFP features \\
& FSIM \cite{zhang2011fsim} & 3 & Phase congruency, gradient magnitude, and chrominance similarity \\
& GMSD \cite{xue2013gradient} & 2 & Mean and standard deviation of gradient magnitude similarity \\
& WS-PSNR \cite{sun2017weighted} & 3 & PSNR of Y, U, V components\\
& CPP-PSNR \cite{Zakharchenko2016} & 3 & PSNR of Y, U, V components\\
& S-PSNR \cite{yu2015framework} & 3 & PSNR of Y, U, V components\\
\hline \multirow{6}{*}{ Audio~ } 
& PEAQ \cite{thiede2000peaq} & 11 & 11 model output variables before the neural network \\
& STOI \cite{taal2011algorithm} & 1 & The complete algorithm \\
& VISQOL \cite{hines2015visqolaudio} & 3 &  Narrowband, wideband, fullband versions of VISOOL \\
& LLR \cite{hu2007evaluation} & 1 & The complete algorithm \\
& SNR \cite{hu2007evaluation} & 1 & The complete algorithm\\
& seg-SNR \cite{hansen1998effective} & 1 & The complete algorithm \\
\bottomrule
\end{tabular}
\end{adjustbox}
\vspace{-18pt}
\end{table}


The performance of SVR fusion methods can be further improved by substituting scores with quality-aware feature vectors $\mathbf{f}_v$ and $\mathbf{f}_a$, which can be either hand-crafted features or extracted features from existing popular AQA and VQA models. In this way, we can better fuse video and audio quality prediction results by fully utilizing the quality features of audio and video, thereby improving the performance of the entire model. This feature-based fusion method can be expressed as:
\begin{equation}
\vspace{-1pt}
Q_{a v}=\textit{SVR}(\mathbf{f}_v, \mathbf{f}_a).
\vspace{-2pt}
\end{equation}
The video and audio quality-aware feature vectors used here are extracted from some existing AQA and VQA models, which are summarized in Table \ref{Features}. 

\vspace{-6pt}
\section{Experiment Validation}
\vspace{-6pt}
\subsection{Evaluation of Single-mode Models}
\begin{figure}[t]
  \centering
  \begin{subfigure}[b]{0.48\textwidth}
    \includegraphics[width=\textwidth]{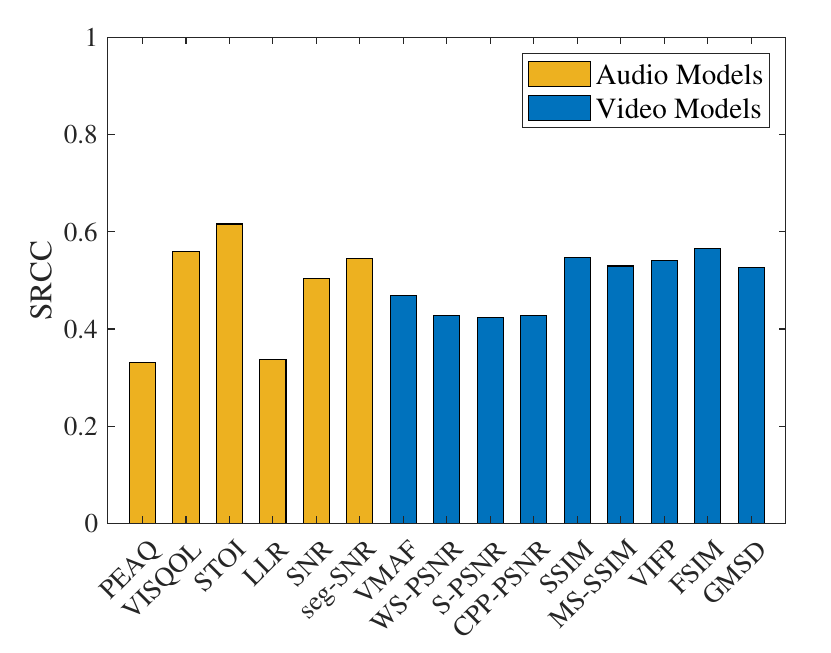}
  \end{subfigure}
  \hspace{0\textwidth}
  \begin{subfigure}[b]{0.48\textwidth}
    \includegraphics[width=\textwidth]{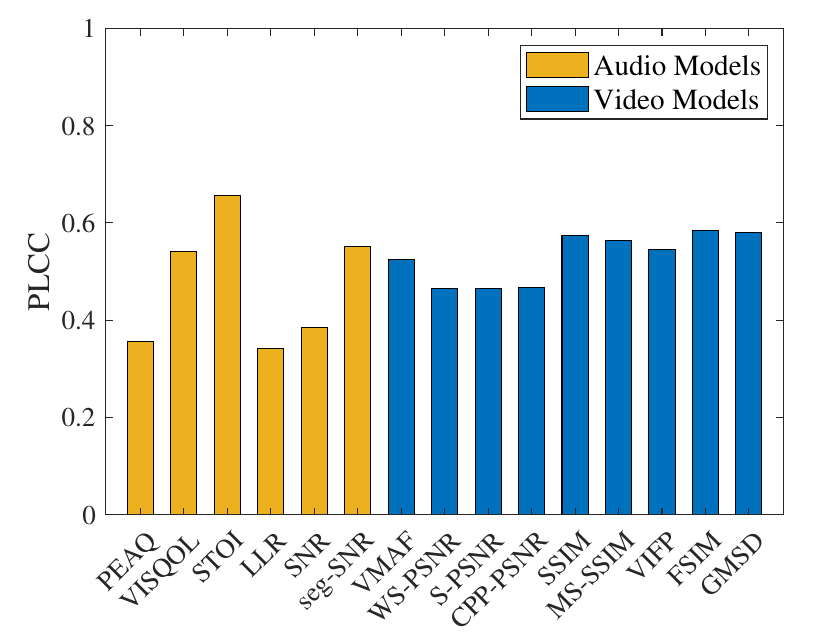}
  \end{subfigure}
  \vspace{-5pt}
  \caption{Performances of single-mode models on overall audio-visual quality prediction.}
  \label{single}
  \vspace{-22pt}
\end{figure}

\vspace{-3pt}
We test different single-mode quality assessment models (6 audio models and 9 video models) on our omnidirectional AVQA dataset to analyze the effectiveness of single-mode quality models. Experimental results are illustrated in Fig.\ref{single}.
 For AQA models, STOI, VISQOL, SNR, and segSNR yield relatively good performances on our database, in which STOI achieves the both highest SRCC and PLCC performance. Most of the VQA models show similar performance, and all of them are not able to predict A/V quality effectively with SRCC and PLCC below 0.6. 
 The above analysis shows that most single-mode quality assessment models have a poor performance on our OAVQAD, indicating the necessity of fusing single-mode quality prediction results for more accurate OAVQA.

\vspace{-8pt}
\subsection{Evaluation of Weighted-product Fusion}
\vspace{-6pt}
For weighted-product fusion methods, we randomly divide the dataset into 80\% training set and 20\% test set. All distorted ODVs from the same reference ODVs are placed in the same set to ensure that the video content of the two set are completely separated. 

In the weighted-product fusion, a total of 54 (9 video models $\times$ 6 audio models) weighted product quality fusion models are generated. In order to normalize the prediction scores of the single-mode quality prediction models, the following normalization functions are used: $ Q_{\textit{VMAF}}^{'}=\frac{1}{100} Q_{\textit{VMAF}}$,   $Q_{\textit{WS-PSNR}}^{'}=\frac{1}{29}(Q_{\textit{WS-PSNR}}-23)$, 
$Q_{\textit{S-PSNR}}^{'}=\frac{1}{29} (Q_{\textit{S-PSNR}}-23)$, 
$Q_{\textit{CPP-PSNR}}^{'}=\frac{1}{29}(Q_{\textit{CPP-PSNR}}-23)$, 
    $Q_{\textit{GMSD}}^{'}=1-\frac{1}{0.26} Q_{\textit{GMSD}}$, 
    $Q_{\textit{PEAQ}}^{'}=1+\frac{1}{3.5} (Q_{\textit{PEAQ}}-0.21)$, 
    $Q_{\textit{LLR}}^{'}=1-\frac{1}{1.2-0.7} (|Q_{\textit{LLR}}|-0.7)$, 
    $Q_{\textit{SNR}}^{'}=\frac{1}{20} Q_{\textit{SNR}}$, 
    $Q_{\textit{segSNR}}^{'}=\frac{1}{35+2} (Q_{\textit{segSNR}}+2)$.
The prediction scores of other models are already bounded in $\left[0,1\right]$, no further normalization is needed.


\begin{table}[!t]
\centering
\setlength{\tabcolsep}{5pt}
\caption{Performances of weighted-product fusion-based A/V quality models. The top 3 models are in bold.}
\label{weighting}
\begin{adjustbox}{width=0.7\textwidth}
\begin{tabular}{|c|c|cccccc|}

\hline \multirow{2}{*}{ Criteria } &  Video  & \multicolumn{6}{c|}{ Weighted Product } \\
\cline { 3 - 8 } & Model & PEAQ & STOI & VISQOL & LLR & SNR & segSNR \\
\hline \multirow{6}{*}{ SRCC } & VMAF & 0.5783 & \textbf{0.7790} & 0.7157 & 0.5745 & 0.7432 & 0.6660 \\
& WS-PSNR & 0.5252 & 0.7348 & 0.6911 & 0.5507 & 0.7124 & 0.6658 \\
& S-PSNR & 0.5182 & 0.7292 & 0.6886 & 0.5460 & 0.7068 & 0.6576 \\
& CPP-PSNR & 0.5246 & 0.7333 & 0.6914 & 0.5499 & 0.7121 & 0.6652 \\
& SSIM & 0.5605 & 0.7717 & 0.7289 & 0.5123 & 0.6783 & 0.6372 \\
& MS-SSIM & 0.6131 & \textbf{0.7998} & 0.7511 & 0.6161 & 0.7596 & 0.6942 \\
& VIFP & 0.5916 & 0.7746 & 0.7332 & 0.5978 & 0.7499 & 0.7017 \\
& FSIM & 0.5386 & 0.7563 & 0.7259 & 0.5638 & 0.6632 & 0.6188 \\
& GMSD & 0.6151 & \textbf{0.8044} & 0.7358 & 0.6246 & 0.7530 & 0.6844 \\
\hline \multirow{6}{*}{ PLCC } & VMAF & 0.6124 & 0.7885 & 0.7265 & 0.6324 & 0.7442 & 0.6484 \\
& WS-PSNR & 0.5595 & 0.7576 & 0.7407 & 0.6020 & 0.7351 & 0.5960 \\
& S-PSNR & 0.5558 & 0.7530 & 0.7404 & 0.5966 & 0.7287 & 0.5984 \\
& CPP-PSNR & 0.5594 & 0.7567 & 0.7401 & 0.6001 & 0.7340 & 0.5931 \\
& SSIM & 0.5984 & 0.7917 & 0.7604 & 0.5601 & 0.6917 & 0.6886 \\
& MS-SSIM & 0.6405 & \textbf{0.8124} & 0.7792 & 0.6561 & 0.7710 & 0.7270 \\
& VIFP & 0.6188 & \textbf{0.8057} & 0.7294 & 0.6415 & 0.7522 & 0.6758 \\
& FSIM & 0.5806 & 0.7743 & 0.7682 & 0.6015 & 0.6693 & 0.6650 \\
& GMSD & 0.6357 & \textbf{0.8112} & 0.7518 & 0.6557 & 0.7587 & 0.6894 \\
\hline
\end{tabular}
\end{adjustbox}
\vspace{-18pt}
\end{table}

\begin{table}[!t]
\centering
\newcolumntype{x}[1]{>{\centering\arraybackslash}p{1.2cm}}
\caption{Performances of SVR fusion-based A/V quality models. The top 3 models in terms of each metric are in bold.}
\label{SVR1}
\begin{adjustbox}{width=0.988\textwidth}
\begin{tabular}
{|c|c|>{\centering\arraybackslash}p{1.2cm}>{\centering\arraybackslash}p{1.2cm}>{\centering\arraybackslash}p{1.2cm}>{\centering\arraybackslash}p{1.2cm}>{\centering\arraybackslash}p{1.2cm}>{\centering\arraybackslash}p{1.2cm}|>{\centering\arraybackslash}p{1.2cm}>{\centering\arraybackslash}p{1.2cm}>{\centering\arraybackslash}p{1.2cm}>{\centering\arraybackslash}p{1.2cm}>{\centering\arraybackslash}p{1.2cm}>{\centering\arraybackslash}p{1.2cm}|}

\hline \multirow{2}{*}{ Criteria } &  Video  & \multicolumn{6}{c|}{ SVR (Quality Score) } & \multicolumn{6}{c|}{ SVR (Quality Feature) }\\
\cline { 3 - 14 } & Model & PEAQ & STOI & ViSQOL & LLR & SNR & segSNR & PEAQ & STOI & VISQOL & LLR & SNR & segSNR\\
\hline \multirow{6}{*}{ SRCC } & VMAF & 0.5481 & 0.7855 & 0.7141 & 0.5676 & 0.5688 & 0.6391 & 0.8343 & 0.8428 & 0.8566 & 0.6052 & 0.6119 & 0.6818\\
& WS-PSNR & 0.5306 & 0.7625 & 0.6974 & 0.5506 & 0.5453 & 0.6269 & 0.8035 & 0.7787 & 0.8171 & 0.5612 & 0.5582 & 0.6346 \\
& S-PSNR & 0.5221 & 0.7593 & 0.6966 & 0.5418 & 0.5365 & 0.6202 & 0.8030 & 0.7764 & 0.8123 & 0.5550 & 0.5476 & 0.6263 \\
& CPP-PSNR & 0.5301 & 0.7626 & 0.6982 & 0.5495 & 0.5452 & 0.6272 & 0.8039 & 0.7806 & 0.8174 & 0.5612 & 0.5584 & 0.6356 \\
& SSIM & 0.5023 & 0.7246 & 0.6734 & 0.4651 & 0.5636 & 0.6222 & 0.7385 & 0.7475 & 0.7654 & 0.5314 & 0.5643 & 0.6492 \\
& MS-SSIM & 0.5809 & 0.7984 & 0.7407 & 0.5963 & 0.6020 & 0.6727 & 0.8201 & 0.8342 & 0.8654 & 0.6136 & 0.6103 & 0.6752 \\
& VIFP & 0.5983 & \textbf{0.8412} & \textbf{0.8149} & 0.6149 & 0.6043 & 0.6887 & \textbf{0.8751} & \textbf{0.8726} & \textbf{0.8881} & 0.6545 & 0.6464 & 0.7311 \\
& FSIM & 0.5046 & 0.7290 & 0.6775 & 0.4604 & 0.5727 & 0.6227 & 0.7485 & 0.7413 & 0.7646 & 0.5275 & 0.5603 & 0.6357 \\
& GMSD & 0.5749 & \textbf{0.8048} & 0.7450 & 0.6178 & 0.6020 & 0.6669 & 0.8426 & 0.7982 & 0.8459 & 0.6084 & 0.5940 & 0.6599 \\
\hline \multirow{6}{*}{ PLCC } & VMAF & 0.5845 & 0.8111 & 0.7808 & 0.6275 & 0.6075 & 0.6832 & 0.8440 & 0.8543 & 0.8619 & 0.6527 & 0.6552 & 0.7303 \\
& WS-PSNR & 0.5789 & 0.7831 & 0.7521 & 0.6125 & 0.5978 & 0.6704 & 0.8113 & 0.8012 & 0.8286 & 0.6312 & 0.6118 & 0.6919 \\
& S-PSNR & 0.5703 & 0.7802 & 0.7472 & 0.6044 & 0.5895 & 0.6599 & 0.8109 & 0.7975 & 0.8229 & 0.6268 & 0.6030 & 0.6800 \\
& CPP-PSNR & 0.5770 & 0.7832 & 0.7514 & 0.6115 & 0.5974 & 0.6712 & 0.8118 & 0.8026 & 0.8286 & 0.6313 & 0.6122 & 0.6930 \\
& SSIM & 0.4297 & 0.7340 & 0.7125 & 0.4892 & 0.4234 & 0.5577 & 0.7656 & 0.7729 & 0.7721 & 0.5641 & 0.5674 & 0.6404 \\
& MS-SSIM & 0.6187 & 0.8168 & 0.7874 & 0.6542 & 0.6476 & 0.7075 & 0.8350 & 0.8508 & 0.8697 & 0.6630 & 0.6632 & 0.7191 \\
& VIFP & 0.6358 & \textbf{0.8565} & \textbf{0.8374} & 0.6752 & 0.6591 & 0.7382 & \textbf{0.8779} & \textbf{0.8828} & \textbf{0.8941} & 0.6950 & 0.6862 & 0.7748 \\
& FSIM & 0.4275 & 0.7330 & 0.7102 & 0.4827 & 0.4298 & 0.5427 & 0.7647 & 0.7647 & 0.7676 & 0.5556 & 0.5595 & 0.6286 \\
& GMSD & 0.6065 & \textbf{0.8249} & 0.7986 & 0.6564 & 0.6495 & 0.6964 & 0.8473 & 0.8170 & 0.8539 & 0.6488 & 0.6409 & 0.6889 \\
\hline
\end{tabular}
\end{adjustbox}\vspace{-20pt}
\end{table}

Table \ref{weighting} shows the performance of weighted product fusion models. Among these methods, the models fused by VQA algorithms VMAF, MS-SSIM, GMSD, and the AQA algorithms STOI, VISQOL, SNR show relatively better performances. The model combining GMSD and STOI achieves the best performance in terms of SRCC. In addition, with the same AQA components, the performance of fusion models using different VQA components has little difference, which manifests that different AQA components have larger impact on the performance of fusion models. 
Moreover, the mean optimal weight for visual modality of 54 weighted product models is 0.7231, suggesting that visual modality has a greater impact on QoE than audio modality.

\vspace{-8pt}
\subsection{Evaluation of SVR Fusion}
\vspace{-6pt}
 SVR fusion includes two methods including the score-based fusion and the feature-based fusion. A total of 108 (9 video models $\times$ 6 audio models $\times$ 2 SVR conditions) models are tested and the normalization process is no longer required. In SVR fusion models, the radial basis function (RBF) is selected as the kernel function, the parameter $\gamma$ of the kernel function is $0.05$, and the penalty factor C is $1024$. Table \ref{SVR1} shows the  performance of SVR fusion models.

It can be observed that quality score-based SVR fusion models achieve similar performance compared with the weighted-product fusion models, while quality feature-based SVR fusion models achieve much better performance compared to above two methods. The models combining the AQA components, PEAQ, STOI and VISQOL, and the VQA components VIFP and GMSD have relatively better performance.

\begin{figure}[!t] 
\vspace{-10pt}
\center{\includegraphics[width=0.63\textwidth]  {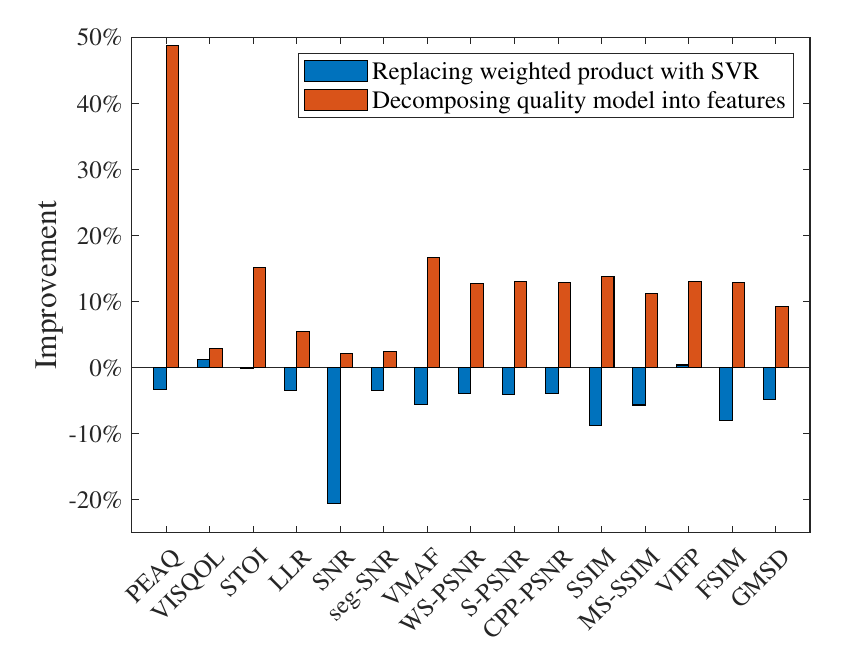}}
\vspace{-8pt}
\caption{Performance improvements in terms of 
SRCC introduced by replacing weighted-product fusion with quality score-based SVR fusion, and decomposing quality models into features during SVR fusion.}
\label{improve}
\vspace{-20pt}
\end{figure}

Fig.\ref{improve} demonstrates the performance improvement obtained by each single-mode AQA and VQA model, which further confirms the above phenomenon. The performance improvement of each single-mode model is calculated by averaging the SRCC improvements of all combinations of this model with the models from another perceptual mode.
It can be observed that only VISQOL and VIFP models gain performance improvement by replacing weighted-product with SVR, suggesting that the weighted-product fusion is generally a more feasible method. Futhermore, Fig.\ref{improve} also illustrates that it is more efficient to decompose the single-mode VQA and AQA scores into ODVs' quality features. It can be observed that the feature-based regression models achieve different degrees of performance improvement for different VQA and AQA fusion, among which PEAQ achieved a significant improvement with nearly 50\%. Some of these models, \textit{e.g.}, STOI, LLR, SNR and segSNR, have a small performance progress caused by feature extraction, we reasonably speculate that these algorithm models are not easy to decompose.

\vspace{-12pt}
\section{Conclusion}
\vspace{-8pt}
In this work, we construct an informative omnidirectional audio-visual quality assessment dataset, which involves 390 omnidirectional videos with ambisonics and the corresponding perceptual scores collected from 22 participants under immersive environment. 
Based on our dataset, we design three types of baseline AVQA models which combine AQA and VQA models via two multimodal fusion methods to predict quality scores of ODVs. Moreover, quantitative analyses for the performance of these models are conducted to evaluate the predictive effect of different objective models. The experiment results on our dataset show that SVR fusion based on quality-aware features have the best performance.
Our dataset, objective baseline methods and established benchmark can great facilitate the further research of dataset design and algorithm improvement for OAVQA.

\vspace{-12pt}
\subsubsection* {Acknowledgement.} This work is supported by National Key R\&D Project of China (2021YFE0206700), NSFC (61831015, 62101325, 62101326, 62271312, 62225112), Shanghai Pujiang Program (22PJ1407400), Shanghai Municipal Science and Technology Major Project (2021SHZDZX0102), STCSM (22DZ2229005).

\bibliographystyle{splncs04}
\bibliography{main.bib}
%




\end{document}